\documentclass{article}

\usepackage{microtype}
\usepackage{graphicx}
\usepackage[pdftex,dvipsnames]{xcolor}  
\usepackage{xargs}                      
\usepackage[colorinlistoftodos,prependcaption,textsize=tiny]{todonotes}
\usepackage{subfigure}
\usepackage{booktabs} 
\usepackage{amsmath}          
\usepackage{amssymb}          
\usepackage[shortlabels]{enumitem}

\usepackage{hyperref}

\usepackage[accepted]{icml2020}

\usepackage{lipsum}

\newcommandx{\unsure}[2][1=]{\todo[linecolor=blue,backgroundcolor=blue!25,bordercolor=blue,#1]{#2}}
\newcommandx{\change}[2][1=]{\todo[linecolor=red,backgroundcolor=red!25,bordercolor=red,#1]{#2}}
\newcommandx{\info}[2][1=]{\todo[linecolor=OliveGreen,backgroundcolor=OliveGreen!25,bordercolor=OliveGreen,#1]{#2}}

\icmltitlerunning{Causal Analysis of Agent Behavior for AI Safety}

\begin{document}

\rfoot{\vspace{20pt} Page \thepage}

\twocolumn[
\icmltitle{Causal Analysis of Agent Behavior for AI Safety}

\icmlsetsymbol{equal}{*}

\begin{icmlauthorlist}
\icmlauthor{Gr\'egoire D\'eletang}{equal,dm}
\icmlauthor{Jordi Grau-Moya}{equal,dm}
\icmlauthor{Miljan Martic}{equal,dm}
\icmlauthor{Tim Genewein}{dm}
\icmlauthor{Tom McGrath}{dm}
\icmlauthor{Vladimir Mikulik}{dm}
\icmlauthor{Markus Kunesch}{dm}
\icmlauthor{Shane Legg}{dm}
\icmlauthor{Pedro A. Ortega}{dm}
\end{icmlauthorlist}

\icmlaffiliation{dm}{AGI Safety Analysis, DeepMind, London, UK}

\icmlcorrespondingauthor{Pedro A. Ortega}{pedroortega@google.com}

\icmlkeywords{Machine Learning, ICML}

\vskip 0.3in
]



\printAffiliationsAndNotice{\icmlEqualContribution} 

\begin{abstract}
As machine learning systems become more powerful they also become increasingly unpredictable and opaque. Yet, finding human-understandable explanations of how they work is essential for their safe deployment. This technical report illustrates a methodology for investigating the  causal mechanisms that drive the behaviour of artificial agents. Six use cases are covered, each addressing a typical question an analyst might ask about an agent. In particular, we show that each question cannot be addressed by pure observation alone, but instead requires conducting experiments with systematically chosen manipulations so as to generate the correct causal evidence.

\end{abstract}

\section{Introduction}

Unlike systems specifically engineered for solving a narrowly-scoped task, machine learning systems such as deep reinforcement learning agents are notoriously opaque. Even though the architecture, algorithms, and training data are known to the designers, the complex interplay between these components gives rise to a black-box behavior that is generally intractable to predict. This problem worsens as the field makes progress and AI agents become more powerful and general. As illustrated by learning-to-learn approaches, learning systems can use their experience to induce algorithms that shape their entire information-processing pipeline, from perception to memorization to action \citep{wang2016learning, andrychowicz2016learning}.

Such poorly-understood systems do not come with the necessary safety guarantees for deployment. From a safety perspective, it is therefore paramount to develop black-box methodologies (e.g.\ suitable for any agent architecture) that allow for investigating and uncovering the causal mechanisms that underlie an agent's behavior. Such methodologies would enable analysts to explain, predict, and preempt failure modes \citep{russell2015research, amodei2016concrete, leike2017ai}. 

This technical report outlines a methodology for investigating agent behavior from a mechanistic point of view. Mechanistic explanations deliver a deeper understanding of agency because they describe the cause-effect relationships that govern behavior---they explain \emph{why} an agent does what it does. Specifically, agent behavior ought to be studied using the tools of causal analysis \citep{spirtes2000causation, pearl2009causality, dawid2015statistical}. In the methodology outlined here, analysts conduct experiments in order to confirm the existence of hypothesized behavioral structures of AI systems. In particular, the methodology encourages proposing simple causal explanations that refer to high-level concepts (``the agent prefers green over red apples'') that abstract away the low-level (neural) inner workings of an agent. 

Using a simulator, analysts can place pre-trained agents into test environments, recording their reactions to various inputs and interventions under controlled experimental conditions. The simulator provides additional flexibility in that it can, among other things, reset the initial state, run a sequence of interactions forward and backward in time, change the seed of the pseudo-random number generator, or spawn a new branch of interactions. The collected data from the simulator can then be analyzed using a causal reasoning engine where researchers can formally express their assumptions by encoding them as causal probabilistic models and then validate their hypotheses. Although labor-intensive, this human-in-the-loop approach to agent analysis has the advantage of producing human-understandable explanations that are mechanistic in nature.


\section{Methodology}

We illustrate this methodology through six use cases, selected so as to cover a spectrum of prototypical questions an agent analyst might ask about the mechanistic drivers of behavior. For each use case, we present a minimalistic grid-world example and describe how we performed our investigation. We limit ourselves to environmental and behavioral manipulations, but direct interventions on the internal state of agents are also possible. The simplicity in our examples is for the sake of clarity only; conceptually, all solution methods carry over to more complex scenarios under appropriate experimental controls.

Our approach uses several components: an agent and an environment, a simulator of interaction trajectories, and a causal reasoning engine. These are described in turn.

\subsection{Agents and environments}\label{sec:agent-env}

\begin{figure}[ht]
\vskip 0.2in
\begin{center}
\centerline{\includegraphics[width=\columnwidth]{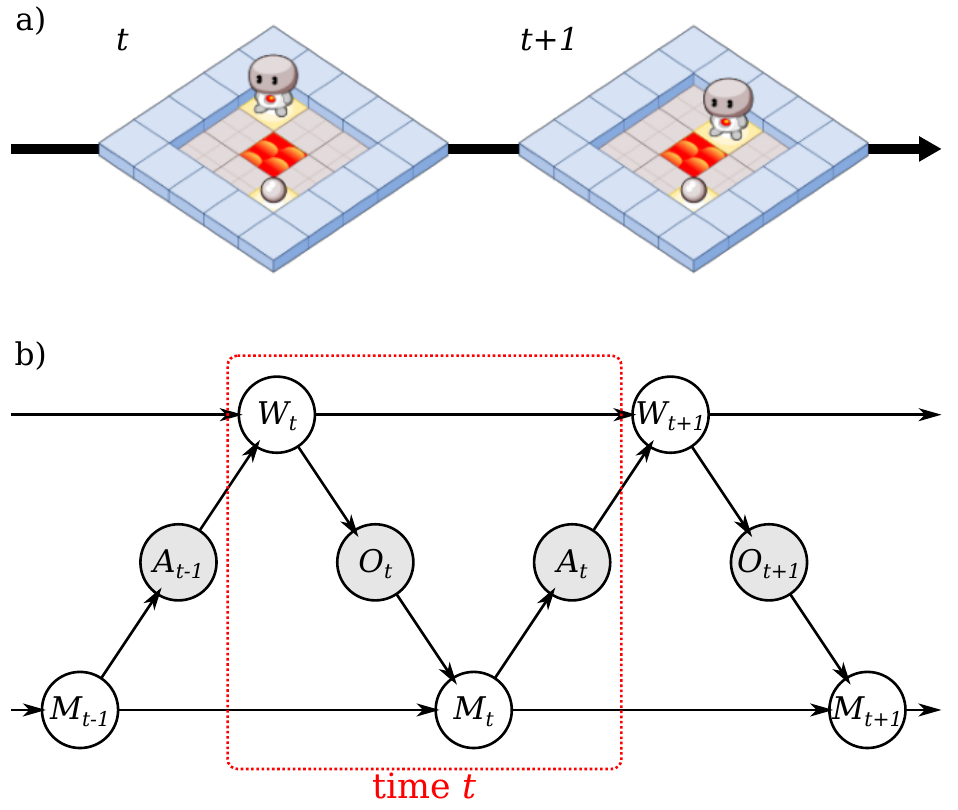}}
\caption{Agents and environments. a)~The goal of the agent is to pick up a reward pill without stepping into a lava tile. b)~Causal Bayesian network describing the generative process of agent-environment interactions. The environmental state~$W_t$ and the agent's memory state~$M_t$ evolve through the exchange of action and observation symbols~$A_t$ and~$O_t$ respectively.}
\label{fig:interaction-diagram}
\end{center}
\vskip -0.2in
\end{figure}

For simplicity, we consider stateful agents and environments that exchange interaction symbols (i.e.\ actions and observations) drawn from finite sets in chronological order at discrete time steps $t=1, 2, 3, \ldots$ Typically, the agent is a system that was pre-trained using reinforcement learning and the environment is a partially-observable Markov decision process, such as in Figure~\ref{fig:interaction-diagram}a. Let $m_t, w_t$ (agent's memory state, world state) and $a_t, o_t$ (action, observation) denote the internal states and interaction symbols at time~$t$ of the agent and the environment respectively. These interactions influence the stochastic evolution of their internal states according to the following (causal) conditional probabilities:
\begin{align}
  w_t &\sim P(w_t \mid w_{t-1}, a_{t-1}) & o_t &\sim P(o_t \mid w_t) \\
  m_t &\sim P(m_t \mid m_{t-1}, o_t) & a_t &\sim P(a_t \mid m_t).
\end{align}
These dependencies are illustrated in the causal Bayesian network of Figure~\ref{fig:interaction-diagram}b describing the perception-action loop \citep{tishby2011information}.

Since we wish to have complete control over the stochastic components of the interaction process (by controlling its random elements), we turn the above into a deterministic system through a re-parameterization\footnote{That is, we describe the system as a structural causal model as described in \citet[][chapter 7]{pearl2009causality}. Although this parameterization is chosen for the sake of concreteness, others are also possible.}. Namely, we represent the above distributions using functions $W, M, O, A$ as follows:
\begin{align}
  w_t &= W(w_{t-1}, a_{t-1}, \omega) & o_t &= O(w_t, \omega) \\
  m_t &= M(m_{t-1}, o_t, \omega) & a_t &= A(m_t, \omega)
\end{align}
where $\omega \sim P(\omega)$ is the random seed. This re-parameterization is natural in the case of agents and environments implemented as programs.

\begin{figure*}[ht]
\vskip 0.2in
\begin{center}
\centerline{\includegraphics[width=0.9\textwidth]{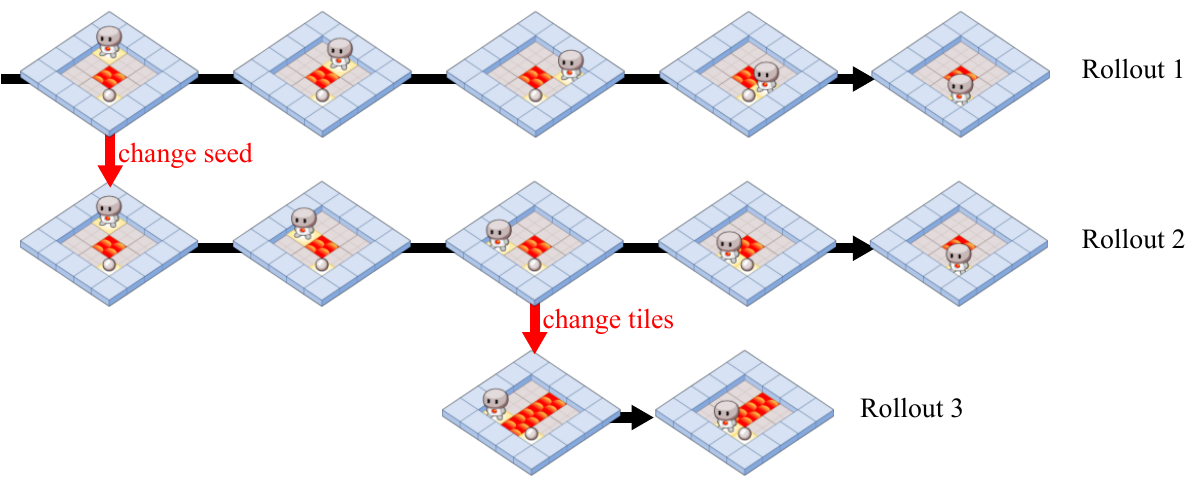}}
\caption{Simulating a trace (rollout) and performing interventions, creating new branches.}
\label{fig:rollouts}
\end{center}
\vskip -0.2in
\end{figure*}

\subsection{Simulator}

The purpose of the simulator is to provide platform for experimentation. Its primary function is to generate traces (rollouts) of agent-environment interactions (Figure~\ref{fig:rollouts}). Given a system made from coupling an agent and an environment, a random seed $\omega \sim P(\omega)$, and a desired length~$T$, it generates a trace
\[
  \tau = (\omega, s_1, x_1),  (\omega, s_2, x_2), (\omega, s_3, x_3), \ldots, (\omega, s_T, x_T)
\]
of a desired length~$T$, where the $s_t:=(\omega_t, m_t)$ and $x_t:=(o_t, a_t)$ are the combined state and interaction symbols respectively, and where~$\omega$ is the random element which has been made explicit. The simulator can also contract (rewind) or expand the trace to an arbitrary time point $T' \geq 1$. Note that this works seamlessly as the generative process of the trace is deterministic.

In addition, the simulator allows for manipulations of the trace. Such an intervention at time~$t$ can alter any of the three components of the triple $(\omega, s_t, x_t)$. For instance, changing the random seed in the first time step corresponds to sampling a new trajectory:
\begin{equation}
\begin{gathered}
  \tau = (\omega, s_1, x_1),  (\omega, s_2, x_2), \ldots, (\omega, s_T, x_T) \\
  \downarrow\\
  \tau' = (\omega', s'_1, x'_1),  (\omega', s'_2, x'_2), \ldots, (\omega', s'_T, x'_T);
\end{gathered}
\end{equation}
whereas changing the state at time step $t=2$ produces a new branch of the process sharing the same root:
\begin{equation}
\begin{gathered}
  \tau = (\omega, s_1, x_1),  (\omega, s_2, x_2), \ldots, (\omega, s_T, x_T) \\
  \downarrow\\
  \tau' = (\omega, s_1, x_1),  (\omega, s'_2, x'_2), \ldots, (\omega, s'_T, x'_T).
\end{gathered}
\end{equation}
Using these primitives one can generate a wealth of data about the behavior of the system. This is illustrated in Figure~\ref{fig:rollouts}.

\subsection{Causal reasoning engine}

Finally, in order to gain a mechanistic understanding of the agent's behavior from the data generated by the simulator, it is necessary to use a formal system for reasoning about statistical causality. The purpose of the causal reasoning engine is to allow analysts to precisely state and validate their causal hypotheses using fully automated deductive reasoning algorithms.

As an illustration of the modeling process, consider an analyst wanting to understand whether an agent avoids lava when trying to reach a goal state. First, the analyst selects the set of random variables $\mathcal{X}$ they want to use to model the situation\footnote{There are some subtleties involved in the selection of random variables. For example, if you want to be able to make arbitrary interventions, the variables should be logically independent. \citet{halpern2011actual} provide a discussion.}. The variables could consist of (abstract) features computed from the trajectories (e.g. ``agent takes left path'') and hypothesis variables (e.g.\ ``the agent avoids lava tiles''). The objective is to obtain a simplified model that abstracts away all but the relevant features of the original interaction system.

\begin{figure}[h!]
\vskip 0.2in
\begin{center}
\centerline{\includegraphics[width=0.6\columnwidth]{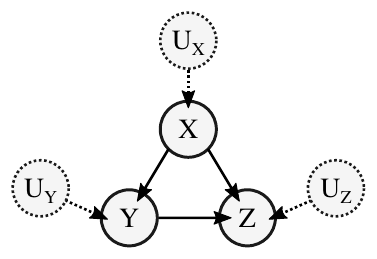}}
\caption{A graphical model representing the structural causal model in \eqref{eq:scm-example}.}
\label{fig:scm-example}
\end{center}
\vskip -0.2in
\end{figure}

Next, the analyst specifies a \emph{structural causal model} \citep[Chapter 7]{pearl2009causality} to describe the causal generative process over the chosen random variables. To illustrate, consider an experiment that can be described using three random variables, $\mathcal{X} = \{X, Y, Z\}$. Assume that $X$ precedes $Y$, and $Y$ in turn precedes $Z$, as shown in Figure~\ref{fig:scm-example}. A structural causal model for this situation would be the system of equations 
\begin{equation}
\begin{aligned}
  X &= f_X(U_X) & U_X &\sim P(U_X) \\
  Y &= f_Y(X, U_Y) & U_Y &\sim P(U_Y) \\
  Z &= f_Z(X, Y, U_Z) & U_Z &\sim P(U_Z)
  \label{eq:scm-example}
\end{aligned}
\end{equation}
where $f_X, f_Y$, and $f_Z$  are (deterministic) functions and where the (exogenous) variables $U_X, U_Y$, $U_Z$ encapsulate the stochastic components of the model. Together, they induce the conditional probabilities
\begin{equation}
  P(X), \quad P(Y \mid X), \quad \text{and} \quad P(Z \mid X, Y).
\end{equation}
These probabilities can be directly supplied by the analyst (e.g.\ if they denote prior probabilities over hypotheses) or estimated from Monte-Carlo samples obtained from the simulator (see next subsection).

Once built, the causal model can be consulted to answer probabilistic queries using the causal reasoning engine. Broadly, the queries come in three types:
\begin{itemize}
 \item \emph{Association:} Here the analyst asks about a conditional probability, such as $P(X=x \mid Y=y)$.
 \item \emph{Intervention:} If instead the analyst controls $Y$ directly, for instance by setting it to the value $Y=y$, then the probability of $X=x$ is given by
 \[
   P(X=x \mid \mathrm{do}(Y=y)).
 \]
 Here, ``$\mathrm{do}$'' denotes the do-operator, which substitutes the equation for $Y$ in the structural model in \eqref{eq:scm-example} with the constant equation $Y=y$. Hence, the new system is
 \begin{equation}
 \begin{aligned}
   X &= f_X(U_X) & U_X &\sim P(U_X) \\
   Y &= y & U_Y &\sim P(U_Y)\\
   Z &= f_Z(X, Y, U_Z) & U_Z &\sim P(U_Z),
   \label{eq:scm-example-do}
 \end{aligned}
 \end{equation}
 which in this case removes the dependency of~$Y$ on~$X$ (and the exogenous variable $U_Y$).
 \item \emph{Counterfactuals:} The analyst can also ask counterfactual questions, i.e.\ the probability of $X=x$ given the event~$Y=y$ had~$Y=y'$ been the case instead. Formally, this corresponds to
 \[
   P(X_y=x \mid Y=y'),
 \] 
 where $X_y$ is the \emph{potential response} of~$X$ when~$Y=y$ is enforced.
\end{itemize}
These correspond to the three levels of the causal hierarchy \cite{pearl2018book}. We refer the reader to \citet{pearl2016causal} for an introduction to causality and \citet{pearl2009causality} for a comprehensive treatment.

\subsection{Analysis workflow}\label{sec:analysis_pipeline}

A typical analysis proceeds as follows.

\paragraph{Exploratory investigation.} The analyst starts by placing a trained agent (provided by an agent trainer) into one or more test environments, and then probing the agent's behavior through interventions using the simulator. This will inform the analyst about the questions to ask and the variables needed to answer them.

\begin{figure}[t]
\vskip 0.2in
\begin{center}
\centerline{\includegraphics[width=0.8\columnwidth]{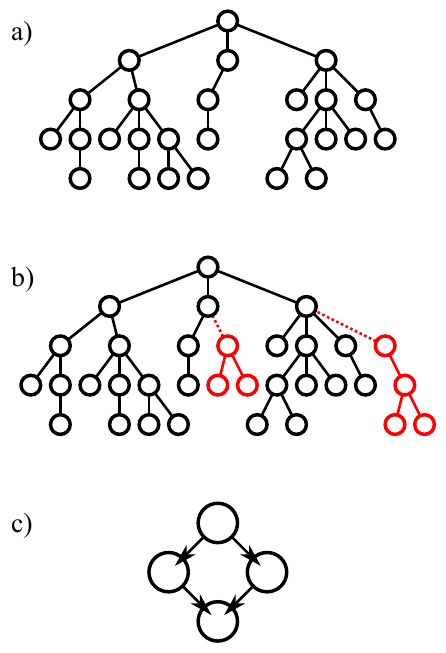}}
\caption{Building a causal model from Monte-Carlo rollouts with interventions. a) A tree generated from Monte-Carlo rollouts from an initial state. This tree contains interaction trajectories that the system can generate by itself. b) When performing experiments, the analyst could enforce transitions (dotted red lines) that the system would never take by itself, such as e.g.\ ``make a lava tile appear next to the agent''. The associated subtrees (red) need to be built from Monte-Carlo rollouts rooted at the states generated through the interventions. c) Finally, the rollout trees can be used to estimate the probabilities of a causal model.}
\label{fig:construction-of-tree}
\end{center}
\vskip -0.2in
\end{figure}

\paragraph{Formulating the causal model.} Next, the analyst formulates a causal model encapsulating all the hypotheses they want to test. If some probabilities in the model are not known, the analyst can estimate them empirically using Monte-Carlo rollouts sampled from the simulator (Figure~\ref{fig:construction-of-tree}a). This could require the use of multiple (stock) agents and environments, especially when the causal hypotheses contrast multiple types of behavior.

In our examples we used discrete random variables. When required, we estimated the conditional probabilities of the causal model following a Bayesian approach. More precisely, for each conditional probability table that had to be estimated, we placed a flat Dirichlet prior over each outcome, and then computed the posterior probabilities using the Monte-Carlo counts generated by the simulator. The accuracy of the estimate can be controlled through the number of samples generated.

Interventions require special treatment (Figure~\ref{fig:construction-of-tree}b). Whenever the analyst performs an intervention that creates a new branch (for instance, because the intervention forces the system to take a transition which has probability zero), the transition probabilities of the subtree must be estimated separately. The transition taken by the intervention itself has zero counts, but it has positive probability mass assigned by the Dirichlet prior. Interventions that do not generate new branches do not require any special treatment as they already have Monte-Carlo samples.

\paragraph{Queries.} Once built (Figure~\ref{fig:construction-of-tree}c), the analyst can query the causal model to answer questions of interest. These can/should then also be verified empirically using the simulator.

\section{Experiments}\label{sec:experiments}

In the following, we present six use cases illustrating typical mechanistic investigations an analyst can carry out:
\begin{itemize}
  \item estimating causal effects under confounding;
  \item testing for the use of internal memory;
  \item measuring robust generalization of behavior;
  \item imagining counterfactual behavior;
  \item discovering causal mechanisms;
  \item and studying the causal pathways in decisions.
\end{itemize}
In each case we assume the agent trainer and the analyst do not share information, i.e.\ we assume the analyst operates under black box conditions. However, the analyst has access to a collection of pre-trained stock agents, which they can consult/use for formulating their hypotheses.

The environments we use were created using the Pycolab game engine \citep{stepletonpycolab}. They are 2D gridworlds where the agent can move in the four cardinal directions and interact with objects through pushing or walking over them. Some of the objects are rewards, doors, keys, floors of different types, etc. The agent's goal is to maximize the sum of discounted cumulative rewards \citep{puterman2014markov, sutton2018reinforcement}. The environments use a random seed for their initialization (e.g.\ for object positions).

In theory, the agents can be arbitrary programs that produce an action given an observation and an internal memory state; but here we used standard deep reinforcement learning agents with a recurrent architecture (see Appendix).


\subsection{Causal effects under confounding}\label{sec:confounding}

\paragraph{Problem.}
Do rewards guide the agent, or do other factors control its behavior? Estimating causal effects is the quintessential problem of causal inference. The issue is that simply observing how the presumed independent and dependent variables co-vary does not suffice, as there could be a third confounding variable creating a spurious association. For instance, sometimes an agent solves a task (e.g.\ picking up a reward pill), but it does so by relying on an accidentally correlated feature (e.g.\ the color of the floor) rather than the intended one (e.g.\ location of the pill). Such policies do not generalize \citep{arjovsky2019invariant}. 

To find out whether the agent has learned the desired causal dependency, one can directly manipulate the independent variable and observe the effect. This manipulation decouples the independent variable from a possible confounder \citep[Chapter 3]{pearl2009causality}. Randomized controlled trials are the classical example of this approach \citep{fisher1936design}.

\begin{figure}[t]
\vskip 0.2in
\begin{center}
\centerline{\includegraphics[width=0.8\columnwidth]{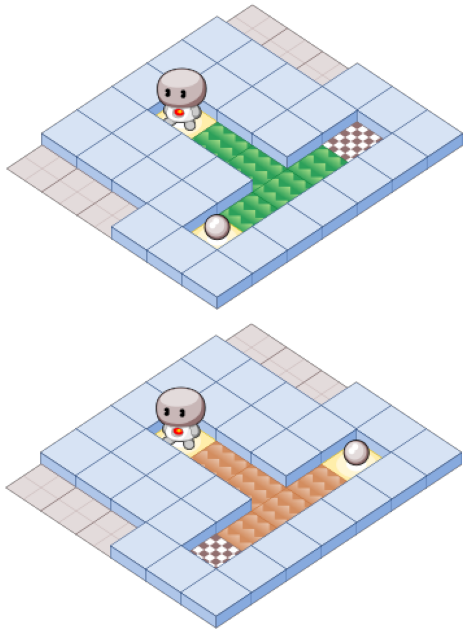}}
\caption{The \emph{grass-sand} environment. The goal of the agent is to pick up a reward pill, located in one of the ends of a T-maze. Reaching either end of the maze terminates the episode. The problem is that the floor type (i.e.\ either grass or sand) is correlated with the location of the reward.}
\label{fig:grass-sand}
\end{center}
\vskip -0.2in
\end{figure}

\paragraph{Setup.}
We illustrate the problem of estimating causal effects using the  \emph{grass-sand} environment depicted in Figure~\ref{fig:grass-sand}. The agent needs to navigate a T-maze in order to collect a pill (which provides a reward) at the end of one of the two corridors \citep{olton1979mazes}. The problem is that the location of the pill (left or right) and the type of the floor (grass or sand) are perfectly correlated. Given an agent that successfully collects the pills, the goal of the analyst is to determine whether it did so because it intended to collect the pills, or whether it is basing its decision on the type of the floor. 

Our experimental subjects are two agents, named A and B. Agent~A was trained to solve T-mazes with either the (sand, left) or (grass, right) configuration; whereas agent~B was trained to solve any of the four combinations of the floor type and reward pill location.

\paragraph{Experiment.}
The experiment proceeds as follows. First, we randomly choose between the (sand, left) and (grass, right) T-mazes and place the agent in the starting position. Then we randomly decide whether to switch the pill location. After this intervention, we let the agent navigate until it finishes the episode, recording whether it took the right or left terminal state.

\begin{figure}[t]
\vskip 0.2in
\begin{center}
\centerline{\includegraphics[width=0.6\columnwidth]{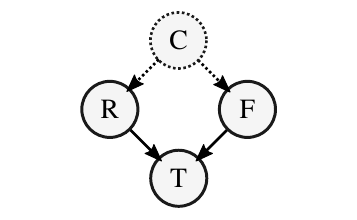}}
\caption{Causal model for the \emph{grass-sand} environment. $R$ is the location of the reward pill; $T$ is the terminal state chosen by the agent; $F$ is the type of the floor; and $C$ is a confounder that correlates $R$ and $F$. Note that $C$ is unobserved.}
\label{fig:grass-sand-model.pdf}
\end{center}
\vskip -0.2in
\end{figure}

We also considered the following hypothesis: namely, that the agent's behavior depends on the type of the floor. To measure the causal effect, we randomly intervened this feature, recording the agent's subsequent choice of the terminal state. The causal model(s) are depicted in Figure~\ref{fig:grass-sand-model.pdf}.

\paragraph{Results.}
Table~\ref{tab:grass-sand} shows the results of the interventions. Here, the random variables $T \in \{l, r\}$, $R \in \{l, r\}$, and $F \in \{g, s\}$ correspond to the agent's choice of the terminal state, the location of the reward pill, and the type of the floor, respectively. The reported values are the posterior probabilities (conditioned on 1000 rollouts) of choosing the left/right terminal for the observational setting (i.e.\ by just observing the behavior of the agent) and for the two interventional regimes. 

The probability of taking the left terminal conditioned on the left placement of the reward was obtained through standard conditioning:
\begin{multline}
  P(T=l \mid R=l) =\\
    \sum_f P(T=l \mid F=f, R=l) P(F=f \mid R=l).
\end{multline}
In contrast, intervening on the reward location required the use of the \emph{adjustment formula} as follows \citep{pearl2009causality}
\begin{multline}
  P(T=l \mid \mathrm{do}(R=l)) =\\
    \sum_f P(T=l \mid F=f, R=l) P(F=f).
\end{multline}
Other quantities were obtained analogously.

We found that the two agents differ significantly. In the observational regime, both agents successfully solve the task, picking up the reward pill. However, manipulating the environmental factors reveals a difference in their behavioral drivers. Agent~A's choice is strongly correlated with the type of floor, but is relatively insensitive to the position of the pill. In contrast, agent~B picks the terminal state with the reward pill, regardless of the floor type.

\begin{table}[h!]
\caption{Grass-sand queries}
\label{tab:grass-sand}
\vskip 0.15in
\begin{center}
\begin{small}
\begin{sc}
\begin{tabular}{lcc}
\toprule
Queries & A & B \\
\midrule
$P(T = l \mid R = l)$ 
  & 0.996 & 0.996 \\
$P(T = r \mid R = r)$ 
  & 0.987 & 0.996 \\
$P(T = l \mid \mathrm{do}(R = l))$ 
  & 0.536 & 0.996 \\
$P(T = r \mid \mathrm{do}(R = r))$ 
  & 0.473 & 0.996 \\
$P(T = l \mid \mathrm{do}(F = g))$ 
  & 0.996 & 0.515 \\
$P(T = r \mid \mathrm{do}(F = s))$ 
  & 0.987 & 0.497 \\
\bottomrule
\end{tabular}
\end{sc}
\end{small}
\end{center}
\vskip -0.1in
\end{table}

\paragraph{Discussion.}
This use case illustrates a major challenge in agent training and analysis: to ensure the agent uses the intended criteria for its decisions. Because it was trained on a collection of environments with a built-in bias, agent~A learned to rely on an undesired, but more salient feature. This is a very common phenomenon. Resolving the use of spurious correlations in learned policies is ongoing research---see for instance \citep{bareinboim2015bandits, arjovsky2019invariant, volodin2020resolving}.

Our experiment shows that inspecting the agent's behavior does not suffice for diagnosing the problem, but independently manipulating the intended decision criterion (i.e.\ the reward location) does. Once the problem is discovered, identifying the confounding factors (e.g.\ the floor type) can be a much harder task for the analyst.


\subsection{Memory}

\paragraph{Problem.}

Does the agent use its internal memory for remembering useful information, or does it off-load the memory onto the environment? Memorization is a necessary skill for solving  complex tasks. It can take place in the agent's internal memory; however, often it is easier for an agent to off-load task-relevant information onto its environment (e.g.\ through position-encoding), effectively using it as an external memory. This difference in strategy is subtle and in fact undetectable without intervening.

To find out whether the agent is actually using its internal memory, we can make mid-trajectory interventions on the environment state variables suspected of encoding task-relevant information. If the agent is using external memory, this will corrupt the agent's decision variables, leading to a faulty behavior. 

\begin{figure}[t!]
\vskip 0.2in
\begin{center}
\centerline{\includegraphics[width=0.9\columnwidth]{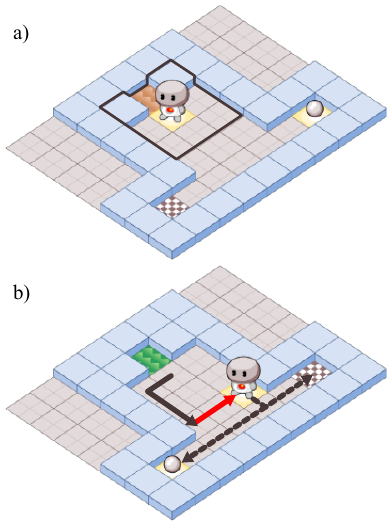}}
\caption{The \emph{floor-memory} environment. \textbf{a)} The goal of the agent with limited vision (see black square) is to collect the reward at one of the ends of the T-maze. A cue informs the agent about the location of the reward. The cue, that can be sand or grass, denotes if the reward is on the right or left, respectively. \textbf{b)} After three steps, we intervene by pushing the agent toward the opposite wall (red arrow), and let it continue thereafter, possibly taking one of the two dashed paths.}
\label{fig:floor-memory}
\end{center}
\vskip -0.2in
\end{figure}

\paragraph{Setup.} We test the agent's memory using the \emph{floor-memory environment} depicted in Figure~\ref{fig:floor-memory}. In this T-maze environment, the agent must remember a cue placed at the beginning of a corridor in order to know which direction to go at the end of it \citep{olton1979mazes, bakker2001reinforcement}. This cue can either be a grass tile or a sand tile, and determines whether the reward is on the right or the left end, respectively. Both cue types and reward locations appear with equal probabilities and are perfectly correlated. The agent can only see one tile around its body.

We consider two subjects. Agent~$a$ is equipped with an internal memory layer (i.e.\ LSTM cells). In contrast, agent~$b$ is implemented as a convolutional neural network without a memory layer; it is therefore unable to memorize any information internally.

\paragraph{Experiment.} 

Gathering rollout data from the test distribution provides no information on whether the agent uses its internal memory or not. An analyst might prematurely conclude that the agent uses internal memory based on observing that the agent consistently solves tasks requiring memorization. However, without intervening, the analyst cannot truly rule out the possibility that the agent is off-loading memory onto the environment.

In this example, we can use the following experimental procedure. First, we let the agent observe the cue and then freely execute its policy. When the agent is near the end of the wide corridor, we intervene by pushing the agent to the opposite wall (see red arrow in Figure~\ref{fig:floor-memory}). This is because we suspect that the agent could use the nearest wall, rather than its internal memory, to guide its navigation. After the intervention, if the agent returns to the original wall and collects the reward, it must be because it is using its internal memory. If on the contrary, the agent does not return and simply continues its course, we can conclude it is off-loading memorization onto its environment.

We model the situation using three random variables. The floor type (grass or sand) is denoted by $F \in \{g, s\}$. The variable $P \in \{l, r\}$ denotes the position of the agent (left or right half-side of the room) at the position when the analyst could execute an intervention. Finally, $T \in \{l, r\}$  represents where the agent is (left or right) when the episode ends. To build the model we randomly decide whether the analyst is going to intervene or not (i.e.\ by pushing) with equal probability. The estimation is performed using $1000$ Monte-Carlo rollouts for each case.

\begin{figure}
    \centering
    \includegraphics[width=0.6\columnwidth]{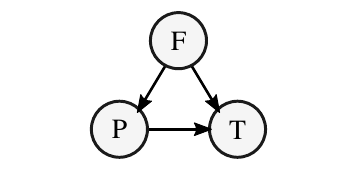}
    \caption{Causal model for the \emph{floor-memory} environment. $F$ is the initial cue (floor type); $P$ is the position of the agent mid-way through the episode; $T$ is the terminal state chosen by the agent. If the agent off-loads the memory about the initial cue onto the position, then the link $F \rightarrow T$ would be missing.}
    \label{fig:trees_floor_memory}
\end{figure}

\paragraph{Results.}  Table~\ref{tab:floor-memory} shows the probabilities obtained by querying the causal model from Figure~\ref{fig:trees_floor_memory}. The first four queries correspond to an observational regime. We see that both agents pick the correct terminal tiles ($T=l$ or $T=r$) with probability close to 1 when conditioning on the cue ($F$) and, additionally, do so by choosing the most direct path ($P=l$ or $P=r$). However, the results from the interventional regime in the last two rows show that agent $A=b$ loses its track when being pushed. This demonstrates that agent~$b$ is using an external memory mechanism that generalizes poorly. In contrast, agent $A=a$ ends up in the correct terminal tile even if it is being pushed to the opposite wall. 

\begin{table}[h!]
\caption{Floor-memory queries for agent $a$ (with internal memory) and $b$ (without internal memory).}
\label{tab:floor-memory}
\vskip 0.15in
\begin{center}
\begin{small}
\begin{sc}
\begin{tabular}{lcc}
\toprule
Queries & $A=a$ & $A=b$ \\
\midrule
$P(T = l \mid F = g)$ & 0.996 & 0.990 \\
$P(T = r \mid F = s)$ & 0.996 & 0.977 \\
$P(P = l \mid F = g)$ & 0.984 & 0.991 \\
$P(P = r \mid F = s)$ & 0.996 & 0.985\\
$P(T = l \mid \mathrm{do}(P = r), F=g)$ & 0.996 & 0.107 \\
$P(T = r \mid \mathrm{do}(P = l), F=s)$ & 0.996 & 0.004 \\
\bottomrule
\end{tabular}
\end{sc}
\end{small}
\end{center}
\vskip -0.1in
\end{table}

\paragraph{Discussion.} 
Agent generalization and performance on partially observable environments depends strongly on the appropriate use of memory. From a safety perspective, flawed memory mechanisms that off-load memorization can lead to fragile behavior or even catastrophic failures. Understanding how AI agents store and recall information is critical to prevent such failures. As shown in the previous experiment, the analyst can reveal the undesired use of external memory by appropriately intervening on the environmental factors that are suspected of being used by the agent to encode task-relevant information.


\subsection{Robust generalization}

\paragraph{Problem.} Does the agent solve any instance within a target class of tasks? Although agents trained through deep reinforcement learning seem to solve surprisingly complex tasks, they struggle to transfer this knowledge to new environments. This weakness is usually hidden by the, unfortunately common, procedure of testing reinforcement learning agents on the same set of environments used for training. Importantly, detecting the failure to generalize to a desired class of environments is key for guaranteeing the robustness of AI agents. 

Two problems arise when assessing the generalization ability of agents. First, testing the agent on the entire class of target environments is typically intractable. Second, the analyst might be interested in identifying the instances within the class of test environments where the agent fails to solve the task, rather than only measuring the average test performance, which could hide the failure modes. This highlights the need for the analyst to assess generalization through the careful choice of multiple targeted tests.

\begin{figure}
    \centering
    \includegraphics[width=0.9\columnwidth]{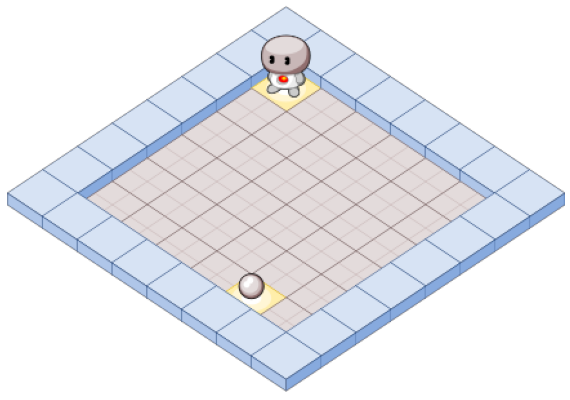}
    \caption{The \emph{pick-up} environment. The goal of the agent is to collect the reward independent of their initial position.}
    \label{fig:apples_full}
\end{figure}

\paragraph{Setup.} 
We illustrate how to test for generalization using the \emph{pick-up} environment shown in Figure~\ref{fig:apples_full}. This is a simple squared room containing a reward which upon collection terminates the episode. The analyst is interested in finding out whether the agent generalizes well to all possible reward locations.

We consider the following two agents as subjects. Both agents were trained on a class of environments where their initial position and the reward location were chosen randomly. However, agent~A's task distribution picks locations anywhere within the room, whereas agent~B's training tasks restricted the location of the reward to the southern quadrant of the room. Thus only agent~A should be general with respect to the class of environments of interest.

\paragraph{Experiment.}
Assume the test set is the restricted class of problem instances where rewards were restricted to the southern corner. Then, if the analyst were to test~A and~B, they could prematurely conclude that both agents generalize. However, assessing generalization requires a different experimental procedure.

The experiment proceeds as follows. We draw an initial state of the system from the test distribution, and subsequently intervene by moving the reward to an arbitrary location within the room. After the intervention, we let the agent freely execute its policy and we observe if the reward was collected or not. A collected reward provides evidence that the agent generalizes under this initial condition. 

\begin{figure}
    \centering
    \includegraphics[width=0.6\columnwidth]{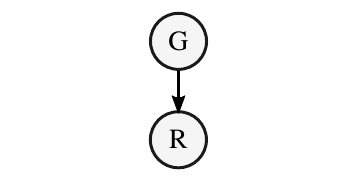}
    \caption{Causal model for the \emph{pick-up} environment. $G$ is the location of the reward pill and $R$ is a binary variable indicating a successful pick-up.}
    \label{fig:tree_agentAandB}
\end{figure}

We built one causal model per agent from $1000$ intervened Monte-Carlo rollouts. The variables are: $G \in \{n, s, e, w\}$, the quadrant location of the reward (north, south, east, west); and $R \in \{0, 1\}$, denoting whether the reward is collected or not. Figure~\ref{fig:tree_agentAandB} shows the causal graph for both models.

\paragraph{Results.}
We performed a number of queries on the causal models shown in Table~\ref{tab:universality}.
Firstly, both agents perform very well when evaluated on the test distribution over problem instances, since $P(R=1) \approx 1$ in both cases. However, the intervened environments tell a different story. As expected, agent~A performs well on all locations of the reward, suggesting that meta-training on the general task distribution was sufficient for acquiring the reward location invariance. Agent~B performs well when the reward is in the southern quadrant, but under-performs in the rest of the conditions. Interestingly, the performance decays as the distance from the southern quadrant increases, suggesting that there was some degree of topological generalization.

\begin{table}[h!]
\caption{Pick-up environment queries for agents $A=a$ and $A=b$.}
\label{tab:universality}
\vskip 0.15in
\begin{center}
\begin{small}
\begin{sc}
\begin{tabular}{lcc}
\toprule
Queries & $A=a$ & $A=b$ \\
\midrule
$P(R = 1)$ & 0.988 & 0.965 \\
$P(R = 1 \mid \mathrm{do}(G=n))$ & 0.985 & 0.230 \\
$P(R = 1 \mid \mathrm{do}(G=e))$ & 0.987 & 0.507 \\
$P(R = 1 \mid \mathrm{do}(G=w))$ & 0.988 & 0.711 \\
$P(R = 1 \mid \mathrm{do}(G=s))$ & 0.988 & 0.986 \\
\bottomrule
\end{tabular}
\end{sc}
\end{small}
\end{center}
\vskip -0.1in
\end{table}

\paragraph{Discussion.}
In this use-case we outlined a procedure for assessing the agents' robust generalization capabilities. Although quantifying generalization in sequential decision-making problems is still an open problem, we adopted a pragmatic approach: we say that an agent generalizes robustly when it successfully completes any task within a desired class of environments. This requirement is related to uniform performance and robustness to adversarial attacks. Since testing all instances in the class is unfeasible, our approximate solution for assessing generalization relies on subdividing the class and estimating the success probabilities within each subdivision. Even if this approximation is crude at the beginning of the analysis, it can provide useful feedback for the analyst. For example, we could further explore agent~B's generalization by increasing the resolution of the reward location.


\subsection{Counterfactuals}

\begin{figure*}[t]
\vskip 0.2in
\begin{center}
\centerline{\includegraphics[width=\textwidth]{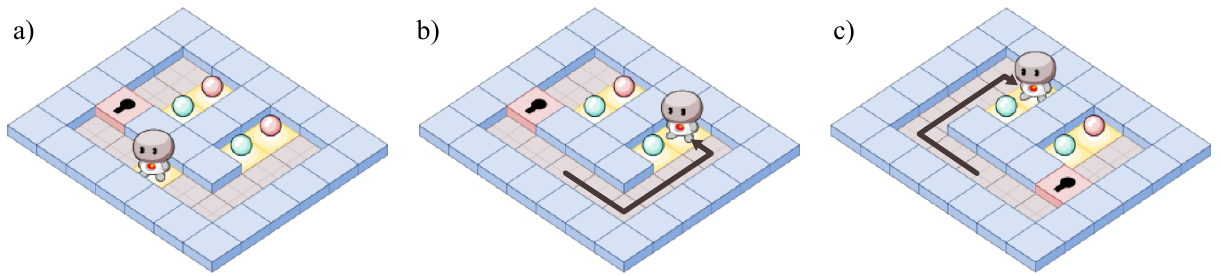}}
\caption{The \emph{gated-room} environments. Panel a: In each instance of the environment, either the left or the right gate will be open randomly. The goal of the agent is to pick up either a red or green reward, after which the episode terminates. Panels b \& c: Counterfactual estimation. If the right door is open and we observe the agent picking up the red reward (b), then we can predict that the agent would pick up the red reward had the left door been open (c).}
\label{fig:gated-room}
\end{center}
\vskip -0.2in
\end{figure*}

\paragraph{Problem.}
What would the agent have done had the setting been different? Counterfactual reasoning is a powerful method assessing an observed course of events. An analyst can imagine changing one or more observed factors without changing others, and imagine the outcome that this change would have led to.

In artificial systems a simulator is often available to the analyst. Using the simulator, the analyst can directly simulate counterfactuals by resetting the system to a desired state, performing the desired change (i.e.\ intervening), and running the interactions ensuing thereafter. This approach yields empirically grounded counterfactuals.

However, simulating counterfactual interactions is not always possible. This happens whenever:
\begin{enumerate}[(a)]    
    \item a realistic simulation for this setting does not exist (e.g.\ for an agent acting in the real world);
    \item a simulation exists, but its use is limited (e.g.\ when evaluating proprietary technology).
\end{enumerate}
For instance, the analyst might be presented with a single behavioral trace of an agent that was trained using an unknown training procedure. Answering counterfactual questions about this agent requires a behavioral model built from prior knowledge about a population of similar or related agents. This is the case which we examine through our experiment. The downside is that such counterfactuals do not make empirically verifiable claims \citep{dawid2000causal}.

\paragraph{Setup.} 
We discuss this problem using the \emph{gated-room environment} depicted in Figure~\ref{fig:gated-room}a. 
The environment consists of two identical rooms each holding a red and a green reward. Collection of the reward terminates the episode. The rooms are initially protected by two gates but one of them randomly opens at the beginning of the episode. We assume there exist two types of agents, classified as either loving green or red reward pills.

\paragraph{Experiment.}
Assume we make a single observation where an unknown agent picks up a red reward in an environment where the right gate is open (Figure~\ref{fig:gated-room}b). We can now ask: ``What would have happened had the left gate been opened instead?'' If we had direct access to the agent's and the environment's internals, we could reset the episode, change which gate is open, and observe what the agent does (Figure~\ref{fig:gated-room}c). But what if this is not possible?

\begin{figure}[ht]
\vskip 0.2in
\begin{center}
\centerline{\includegraphics[width=0.6\columnwidth]{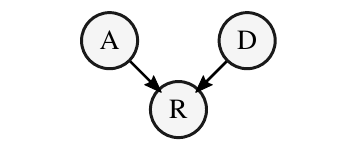}}
\caption{Causal model for the \emph{gated-room} environment. $A$ corresponds to the type of agent (green- or red-pill loving); $D$ indicates which one of the two doors is open; and~$R$ denotes the color of the pill picked up by the agent.}
\label{fig:counterfactual-tree}
\end{center}
\vskip -0.2in
\end{figure}

In order to answer this question, we built a behavioral model using prior knowledge and data. First, we trained two agents that were rewarded for collecting either a green or red reward respectively. These agents were then used to create likelihood models for the two hypotheses using Monte-Carlo sampling. Second, we placed a uniform prior over the two hypotheses and on the open door, and assumed that neither variable precedes the other causally. The resulting causal model, shown in Figure~\ref{fig:counterfactual-tree}, uses three random variables: $A \in \{gr, re\}$ denotes the agent type (green-loving or red-loving); $D \in \{l, r\}$ stands for the open door; and finally $R \in \{gr, re\}$ corresponds to the reward collected by the agent.

\paragraph{Results.}
We performed a number of queries on the model. The results are shown in Table~\ref{tab:counterfactual}. We first performed three sanity checks. Before seeing any evidence, we see that the prior probabilities $P(R=gr)$ and $P(R=re)$ of a random agent picking either a green or a red reward is $0.5$. After observing the agent picking up a red reward ($R=re$) when the left gate is open ($D=l$), we conclude that it must be a red-loving agent ($A=re$) with probability $0.9960$. Note that since the hypothesis about the agent type and the opened door are independent, this probability is the same if we remove the door from the condition. 

Having seen a trajectory, we can condition our model and ask the counterfactual question. Formally, this question is stated as
\[
  P(R_{D=r} = re \mid D=l, R=re),
\]
that is, given that we have observed $D=l$ and $R=re$, what is the probability of the potential response $R_{D=r} = re$, that is, $R=re$ had $D=r$ been the case? The result, $0.9920 \approx 1$, tells us that the agent would also have picked up the red reward had the other door been open, which is in line with our expectations. Furthermore, due to the symmetry of the model, we get the same result for the probability of picking a green reward had the right door been open for an agent that picks up a green reward when the left door is open.

\begin{table}[h!]
\caption{Gated-room queries}
\label{tab:counterfactual}
\vskip 0.15in
\begin{center}
\begin{small}
\begin{sc}
\begin{tabular}{lc}
\toprule
Queries & Probability \\
\midrule
$P(R = re)$ & 0.500 \\
$P(A = re \mid R=re)$ & 0.996 \\
$P(A = re \mid D=l, R=re)$ & 0.996 \\
$P(R_{D=r} = re \mid D=l, R=re)$ & 0.992 \\
$P(R_{D=r} = gr \mid D=l, R=gr)$ & 0.992 \\
\bottomrule
\end{tabular}
\end{sc}
\end{small}
\end{center}
\vskip -0.1in
\end{table}

\paragraph{Discussion.} 
Following the example above, we can naturally see that we are only able to ask counterfactual questions about the behavior of a particular agent when we can rely on prior knowledge about a reference agent population. For instance, this is the case when the agent under study was drawn from a distribution of agents for which we have some previous data or reasonable priors. If we do not have a suitable reference class, then we cannot hope to make meaningful counterfactual claims.


\subsection{Causal induction}\label{sec:causal-induction}

\paragraph{Problem.}
What is the causal mechanism driving an observed behavior? Discovering the mechanisms which underlie an agent's behavior can be considered the fundamental problem of agent analysis. All the use cases reviewed so far depend on the analyst knowing the causal structure governing the agent's behavior. However this model is often not available in a black-box scenario. In this case, the first task of the analyst is to discover the behavioral mechanisms through carefully probing the agent with a variety of inputs and recording their responses \citep{griffiths2005structure}.

Discovering causal structure is an induction problem. This is unlike a deduction task, where the analyst can derive unequivocal conclusions from a set of facts. Rather, induction problems do not have right or wrong answers and require maintaining multiple plausible explanations \citep{rathmanner2011philosophical}.

In this use case, we demonstrate how to induce a distribution over competing causal models for explaining an agent's behavior given experimental data. Although temporal order is often informative about the causal dependencies among random variables, the careful analyst must consider the possibility that a cause and its effect might be observed simultaneously or in reversed temporal order. Thus, in general, observing does not suffice: to test a causal dependency the analyst must manipulate one variable and check whether it influences another\footnote{Although, there are cases where partial structure can be deduced from observation alone---see \citet[][Chapter 2]{pearl2009causality}}. This principle is often paraphrased as ``no causes in, no causes out'' \citep{cartwright1994nature}.

\begin{figure}
    \centering
    \includegraphics[width=0.8\columnwidth]{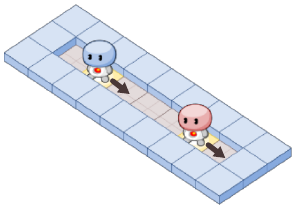}
    \caption{The \emph{mimic} environment. Both agents either step to the left or the right together. The analyst's goal is to discover which one is the lead, and which one is the imitator.}
    \label{fig:mimic}
\end{figure}

\paragraph{Setup.}
We exemplify how to induce a causal dependency using the \emph{mimic} environment shown in Figure~\ref{fig:mimic}. Two agents, blue and red, are placed in a corridor. Then, both agents move simultaneously one step in either direction. One of the two agents is the leader and the other the imitator: the leader chooses its direction randomly, whereas the imitator attempts to match the leader's choice in the same time step, but sampling a random action 10\% of the time. The analyst's task is to find out which agent is the leader. Note there is no way to answer this question from observation alone.

\begin{figure}
    \centering
    \includegraphics[width=0.6\columnwidth]{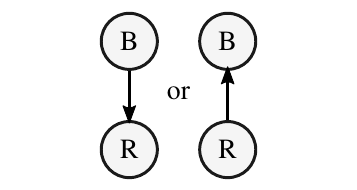}
    \caption{Causal models for the \emph{mimic} environment. Each model has the same prior probability is being correct. $B$ and $R$ indicate the direction in which the blue and the red agents respectively move.}
    \label{fig:tree-causal-induction}
\end{figure}

\paragraph{Experiment.}
We built the causal model as follows. First, we decided to model this situation using three random variables: $L \in \{b, r\}$, corresponding to the hypothesis that either the blue or red agent is the leader, respectively; $B \in \{l, r\}$, denoting the step the blue agent takes; and similarly $R \in \{l, r\}$ for the red agent. The likelihood models were estimated from 1000 Monte-Carlo rollouts, where each rollout consists of an initial and second time step. With the constructed dataset we were able to estimate the joint distribution $P(B, R)$. Since this distribution is purely observational and thus devoid of causal information, we further factorized it according to our two causal hypotheses, namely
\begin{equation}\label{eq:blue-leader}
  P(B, R) = P(B) P(R|B)
\end{equation}
for the hypothesis that blue is the leader ($L=b$), and
\begin{equation}\label{eq:red-leader}
  P(B, R) = P(R) P(B|R)
\end{equation}
for the competing hypothesis ($L=r$). This yields two causal models. Finally, we placed a uniform prior over the two causal models $L=b$ and $L=a$. See Figure~\ref{fig:tree-causal-induction}. Notice that both causal models are observationally indistinguishable.

This symmetry can be broken through intervention. To do so, we force the red agent into a random direction (say, left) and record the response of the blue agent (left). The posterior probabilities over the intervened hypotheses are then proportional to
\begin{align}\label{eq:causal-posterior-b}
  \nonumber
  P(L=b \mid \mathrm{do}&(R=l), B=l) \propto\\
  \nonumber
    &P(L=b) P(B=l|L=b),\quad\text{and}\\
  \nonumber
  P(L=r \mid \mathrm{do}&(R=l), B=l) \propto\\
    &P(L=r) P(B=l|L=r, R=l).
\end{align}
Notice how the intervened factors drop out of the likelihood term.

\begin{table}[h!]
\caption{Mimic queries}
\label{tab:causal-induction}
\vskip 0.15in
\begin{center}
\begin{small}
\begin{sc}
\begin{tabular}{lc}
\toprule
Queries & Probability \\
\midrule
$P(L = b)$ & 0.500 \\
$P(L = b \mid R=l, B=l)$ & 0.500 \\
$P(L = b \mid R=l, B=r)$ & 0.500 \\
$P(L = b \mid \mathrm{do}(R=l), B=l)$ & 0.361 \\
$P(L = b \mid \mathrm{do}(R=l), B=r)$ & 0.823 \\
\bottomrule
\end{tabular}
\end{sc}
\end{small}
\end{center}
\vskip -0.1in
\end{table}

\paragraph{Result.}
We performed the queries shown in Table~\ref{tab:causal-induction}. The first three queries show that observation does not yield evidence for any of the causal hypotheses:
\begin{align*}
  P(L=b) 
  &= P(L=b \mid R=l, B=l) \\
  &= P(L=b \mid R=l, B=r).
\end{align*}
However, pushing the red agent to the left renders the two hypotheses asymmetrical, as can be seen by
\begin{align*}
  P(L=b) 
  &\neq P(L=b \mid \mathrm{do}(R=l), B=l) \\
  &\neq P(L=b \mid \mathrm{do}(R=l), B=r).
\end{align*}
Thus, observing that the blue agent moves to the right after our intervention allows us to conclude that the blue agent is likely to be the leader.

\paragraph{Discussion.}
Our experiment illustrates a Bayesian procedure for discovering the causal mechanisms in agents. The main take-away is that inducing causal mechanisms requires: (a) postulating a collection of causal hypotheses, each one proposing alternative mechanistic explanations for the same observed behavior; and (b) carefully selecting and applying manipulations in order to render the likelihood of observations unequal.


\subsection{Causal pathways}
\paragraph{Problem.} How do we identify an agent's decision-making pathways? In previous examples we have focused on studying how environmental factors influence the agent's behavior. However, we did not isolate the specific chain of mechanisms that trigger a decision. Understanding these pathways is crucial for identifying the sources of malfunction. To estimate the effect of a given pathway, one can chain together the effects of the individual mechanisms along the path \citep{shpitser2013counterfactual, chiappa2019path}.

\paragraph{Setup.}

\begin{figure}
    \centering
    \includegraphics[width=0.8\columnwidth]{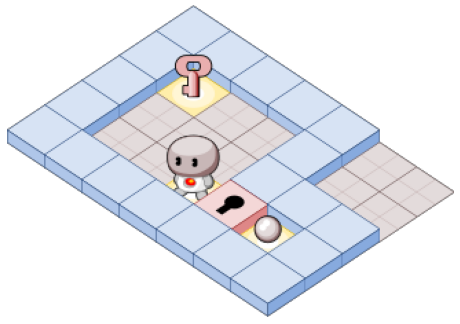}
    \caption{The \emph{key-door} environment. The goal of the agent is to collect the reward, which terminates the episode. However, the reward is behind a door which is sometimes closed. To open it, the agent must collect a key first.}
    \label{fig:key_door_env}
\end{figure}

We illustrate the analysis of causal pathways using the \emph{key-door} environment shown in Figure~\ref{fig:key_door_env}. The agent finds itself in a room where there is a key and a door. The starting position of the agent, the location of the key, and the state of the door (open/closed) are all randomly initialized. Behind the door there is a reward which terminates the episode when picked up. 

We consider two agent subjects. Agent~A appears to only pick-up the key if the door is closed and then collects the reward. This agent acquired this policy by training it on the entire set of initial configurations (i.e. open/closed doors, key and agent positions). Agent~B always collects the key, irrespective of the state of the door, before navigating toward the reward. This behavior was obtained by training the agent only on the subset of instances where the door was closed. Nonetheless, both policies  generalize. The analyst's task is to determine the information pathway used by the agents in order to solve the task; in particular, whether the agent is sensitive to whether the door is open or closed.

\begin{figure}
  \centering
  \includegraphics[width=0.7\columnwidth]{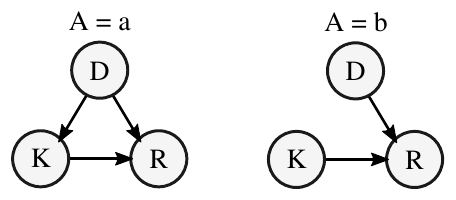}
  \caption{Causal models for the \emph{key-door} environment. $D$ indicates whether the door is open; $K$ flags whether the agent picks up the key; and $R$ denotes whether the agent collects the reward pill. Here, the second model does not include the pathway $D \rightarrow K \rightarrow R$; hence, the agent picks up the key irrespective of the state of the door.}
  \label{fig:key_door_trees}
\end{figure}

\paragraph{Experiment.} 
We chose three random variables to model this situation: $D \in \{o, c\}$, determining whether the door is initially open or closed; $K \in \{y, n\}$, denoting whether the agent picked up the key; and finally, $R \in \{1, 0\}$, the obtained reward. Figure~\ref{fig:key_door_trees} shows the causal models.

\paragraph{Results.}
We investigate the causal pathways through a number of queries listed in Table~\ref{tab:key-door}. First, we verify that both agents successfully solve the task, i.e.\ $P(R=1) \approx 1$.

Now we proceed to test for the causal effect of the initial state of the door on the reward, via the key collection activity. In other words, we want to verify whether $D \rightarrow K \rightarrow R$. This is done in a backwards fashion by chaining the causal effects along a path.

First, we inspect the link $K \rightarrow R$. In the case of agent~A, the reward appears to be independent of whether the key is collected, since
\[
  P(R = 1 \mid K=y) \approx
  P(R = 1 \mid K=n) \approx 1.
\]
However, this is association and not causation. The causal effect of collecting the key is tested by comparing the interventions, that is,
\[
  P(R=1 \mid \mathrm{do}(K=y)) - P(R=1 \mid \mathrm{do}(K=n)).
\]
Here it is clearly seen that both agents use this mechanism for solving the task, since the difference in probabilities is high. This establishes $K \rightarrow R$.

Second, we ask for the causal effect of the initial state of the door on collecting the key, i.e.\ $D \rightarrow K$. Using the same rationale as before, this is verified by comparing the intervened probabilities:
\[
  P(K=y \mid \mathrm{do}(D=c)) - P(K=y \mid \mathrm{do}(D=o)).
\]
Here we observe a discrepancy: agent~A is sensitive to $D$ but agent~B is not. For the latter, we conclude $D \not\rightarrow K \rightarrow R$.

\begin{table}[t]
\caption{Key-door queries}
\label{tab:key-door}
\vskip 0.15in
\begin{center}
\begin{small}
\begin{sc}
\begin{tabular}{lcc}
\toprule
Queries & $A=a$ & $A=b$ \\
\midrule
$P(R = 1)$ & 0.977 & 0.991 \\
--- \\
$P(R = 1 \mid K=y)$ & 0.974 & 0.993 \\
$P(R = 1 \mid K=n)$ & 0.989 & 0.445 \\
$P(R = 1 \mid \mathrm{do}(K=y))$ & 0.979 & 0.993 \\
$P(R = 1 \mid \mathrm{do}(K=n))$ & 0.497 & 0.334 \\
--- \\
$P(K = y \mid \mathrm{do}(D=c))$ & 0.998 & 0.998 \\
$P(K = y \mid \mathrm{do}(D=o))$ & 0.513 & 0.996 \\
---  \\
$P(R = 1 \mid D=c)$ & 0.960 & 0.988 \\
$P(R = 1 \mid D=o)$ & 0.995 & 0.995 \\
$f(D=c)$, see \eqref{eq:npr} & 0.978 & 0.992 \\
$P(D=o)$, see \eqref{eq:npr} & 0.744 & 0.991 \\
\bottomrule
\end{tabular}
\end{sc}
\end{small}
\end{center}
\vskip -0.1in
\end{table}

Finally, we estimate the causal effect the state of the door has on the reward, along the causal pathways going through the settings of $K$. Let us inspect the case $D=c$. The conditional probability is
\begin{align*}
  P&(R=1 \mid D=c) =\\
  &\sum_{k \in \{y,n\}}
    P(R=1 \mid K=k, D=c)
    P(K=k \mid D=c),
\end{align*}
and we can easily verify that $P(R=1 \mid D) \approx P(R=1)$, that is, $D$ and $R$ are independent. But here again, this is just association. The causal response along the pathways is given by
\begin{align}
  \nonumber
  f&(D=c) :=\\
  \label{eq:npr}
  &\sum_{k \in \{y,n\}}
    P(R=1 \mid \mathrm{do}(K=k))
    P(K=k \mid \mathrm{do}(D=c)),
\end{align}
which is known as a \emph{nested potential response} \citep{carey2020incentives} or a \emph{path-specific counterfactual} \citep{chiappa2019path}. The desired causal effect is then computed as the difference between closing and opening the door, i.e.\ 
\[
  f(D=c) - f(D=o).
\]
This difference amounts to 0.2338 and $0.0014 \approx 0$ for the agents~A and~B respectively, implying that~A does indeed use the causal pathway $D \rightarrow K \rightarrow R$ but agent~B only uses $K \rightarrow R$.

\paragraph{Discussion.}
Understanding causal pathways is crucial whenever not only the final decision, but also the specific causal pathways an agent uses in order to arrive at said decision matters. This understanding is critical for identifying the sources of malfunctions and in applications that are sensitive to the employed decision procedure, such as e.g.\ in fairness \citep{chiappa2019path}. In this experiment we have shown how to compute causal effects along a desired path using nested potential responses computed from chaining together causal effects.


\section{Discussion and Conclusions}

\paragraph{Related work.}
The analysis of black-box behavior dates back to the beginnings of electronic circuit theory \citep{cauer1954theorie} and was first formalized in cybernetics \citep{wiener1948cybernetics,ashby1961introduction}, which stressed the importance of manipulations in order to investigate the mechanisms of cybernetic systems. However, the formal machinery for reasoning about causal manipulations and their relation to statistical evidence is a relatively recent development \citep{spirtes2000causation, pearl2009causality, dawid2015statistical}. 

A recent line of research related to ours that explicitly uses causal tools for analyzing agent behavior is \citet{everitt2019understanding} and \citet{carey2020incentives}. These studies use causal incentive diagrams to reason about the causal pathways of decisions in the service of maximizing utility functions.  Other recent approaches for analyzing AI systems have mostly focused on white-box approaches for improving understanding \citep[see for instance][]{mott2019towards, verma2018programmatically, montavon2018methods, explainableRL2020} and developing safety guarantees \cite{uesato2018rigorous}. A notable exception is the work by \citet{rabinowitz2018machine}, in which a model is trained in order to predict agent behavior from observation in a black-box setting.

\paragraph{Scope.}
In this report we have focused on the black-box study of agents interacting with (artificial) environments, but the methodology works in a variety of other settings: passive agents like sequence predictors, systems with interactive user interfaces such as language models and speech synthesizers, and multi-agent systems. For example, consider GPT-3 \cite{brown2020language}, a natural language model with text-based input-output. This system can be seen as a perception-action system, for which our methodology applies. A bigger challenge when dealing with models systems might be to come up with the right hypotheses, problem abstractions, and interventions.

\paragraph{Features and limitations.}
The main challenge in the practice of the proposed methodology is to come up with the right hypotheses and experiments. This task requires ingenuity and can be very labor-intensive (Section~\ref{sec:analysis_pipeline}). For instance, while in the grass-sand environment it was easy to visually spot the confounding variable (Section~\ref{sec:confounding}), we cannot expect this to be a viable approach in general. Or, as we have seen in the problem of causal induction (Section~\ref{sec:causal-induction}), it is non-trivial to propose a model having a causal ordering of the variables that differs from the sequence in which they appear in a sampled trajectory. Given the inherent complexity of reasoning about causal dependencies and the state of the art in machine learning, it is unclear how to scale this process through e.g.\ automation. 

On the plus side, the methodology naturally leads to human-explainable theories of agent behavior, as it is human analysts who propose and validate them. As illustrated in our examples, the explanations do not make reference to the true underlying mechanisms of agents (e.g.\ the individual neuronal activations), but instead rely on simplified concepts (i.e.\ the model variables) that abstract away from the implementation details. See also \citet{rabinowitz2018machine} for a discussion. The human analyst may also choose an appropriate level of detail of an explanation, for instance proposing general models for describing the overall behavior of an agent and several more detailed models to cover the behavior in specific cases.

We have not addressed the problem of quantifying the uncertainty in our models. When estimating the conditional probabilities of the causal models from a limited amount of Monte-Carlo samples, there exists the possibility that these deviate significantly from the true probabilities. In some cases, this could lead to the underestimation of the probability of failure modes. To quantify the reliability of estimates, one should supplement them with confidence intervals, ideally in a manner to aid the assessment of risk factors. In this work we have simply reported the number of samples used for estimation. Developing a more systematic approach is left for future work.

\paragraph{Conclusions and outlook.}
This technical report lays out a methodology for the systematic analysis of agent behavior. This was motivated by experience: previously, we have all too often fallen into the pitfalls of misinterpreting agent behavior due to the lack of a rigorous method in our approach. Just as we expect new medical treatments to have undergone a rigorous causal study, so too do we want AI systems to have been subjected to similarly stringent tests. We have shown in six simple situations how an analyst can propose and validate theories about agent behaviour through a systematic process of explicitly formulating causal hypotheses, conducting experiments with carefully chosen manipulations, and confirming the predictions made by the resulting causal models. Crucially, we stress that this mechanistic knowledge could only be obtained via directly interacting with the system through interventions. In addition, we greatly benefited from the aid of an automated causal reasoning engine, as interpreting causal evidence turns out to be a remarkably difficult task. We believe this is the way forward for analyzing and establishing safety guarantees as AI agents become more complex and powerful. 

\section*{Acknowledgements}

The authors thank Tom Everitt, Jane X. Wang, Tom Schaul, and Silvia Chiappa for proof-reading and providing numerous comments for improving the manuscript. 


\appendix
\section{Architecture and training details}\label{sec:arch-details}
In our experiments we use agents with the following architecture: 3 convolutional layers with 128 channels (for each tile type) each and $3 \times 3$ kernels; a dense linear layer with 128 units; a single LSTM layer with 128 units \citep{hochreiter1997long}; a dense linear layer with 128 units; and a softmax activation layer for producing stochastic actions. To train them, we used the Impala policy gradient algorithm \citep{espeholt2018impala}. The gradients of the recurrent network were computed with backpropagation through time \citep{robinson1987utility, werbos1988generalization}, and we used Adam for optimization \citep{kingma2014adam}. During training, we randomized the environment and agent seed, forcing the agent to interact with different settings and possibly meta-learn a general policy.


\bibliography{bibliography}
\bibliographystyle{icml2020}
\end{document}